\pdfoutput=1

\documentclass[11pt]{article}

\usepackage{booktabs, tabularx}

\usepackage{ACL2023}

\usepackage{times}
\usepackage{latexsym}
\usepackage{amsmath,amssymb,amsfonts}
\usepackage{multirow}
\usepackage{colortbl}

\usepackage[T1]{fontenc}

\usepackage[utf8]{inputenc}

\usepackage{microtype}

\usepackage{inconsolata}

\newcommand{\minus}{$-$}
\newcommand{\plus}{$+$}
\definecolor{ivory2}{RGB}{212,212,212}

\title{NusaBERT: Teaching IndoBERT to be Multilingual and Multicultural}

\author{Wilson Wongso$^{\dagger}$, David Samuel Setiawan$^{\dagger}$, Steven Limcorn$^{\dagger}$, Ananto Joyoadikusumo$^{\dagger}$ \\
        LazarusNLP \\ $^{\dagger}$ Equal Contribution}

\begin{document}
\maketitle
\begin{abstract}
Indonesia's linguistic landscape is remarkably diverse, encompassing over 700 languages and dialects, making it one of the world's most linguistically rich nations. This diversity, coupled with the widespread practice of code-switching and the presence of low-resource regional languages, presents unique challenges for modern pre-trained language models. In response to these challenges, we developed NusaBERT, building upon IndoBERT by incorporating vocabulary expansion and leveraging a diverse multilingual corpus that includes regional languages and dialects. Through rigorous evaluation across a range of benchmarks, NusaBERT demonstrates state-of-the-art performance in tasks involving multiple languages of Indonesia, paving the way for future natural language understanding research for under-represented languages.
\end{abstract}

\section{Introduction}

Indonesia’s exceptional linguistic landscape, encompassing over 700 languages and dialects \cite{aji-etal-2022-one}, presents a significant challenge for current natural language processing (NLP) techniques, such as pre-trained language models. These techniques often fall short in handling the nation’s intricate and multifaceted linguistic tapestry. Furthermore, the bilingual nature of Indonesian colloquial conversations (mixing Indonesian and English) with the majority continuing to also communicate in regional languages as their daily conversational language poses a complex problem to be solved by language models.

Nonetheless, pre-trained language models have shown remarkable progress in recent years showing their ability to solve a wide range of natural language processing tasks, including the Indonesian language. These language models are trained on a large corpus and are fine-tuned to solve specific, downstream tasks. Language models such as BERT \cite{devlin2018bert} and GPT \cite{radford2018improving,radford2019language} are typically trained on a monolingual corpus and were originally trained on an English corpus. In the studies that followed, language-specific language models like IndoBERT \cite{wilie-etal-2020-indonlu} and IndoBART \cite{cahyawijaya-etal-2021-indonlg} have been tailored for the Indonesian language and regional languages of Indonesia like Javanese and Sundanese. Despite the large size discrepancy between the English and Indonesian corpus, IndoBERT managed to leverage the contextualized Indonesian language model to attain exceptional results on multiple downstream natural language understanding (NLU) tasks called IndoNLU.

Although demonstrating remarkable capabilities across various tasks, these pre-trained language models often suffer when applied to languages with unique characteristics like those found in the many regions of Indonesia. For instance, while IndoBERT achieved great results on Indonesian downstream tasks, it still faces limitations when addressing the intricacies of code-switching \cite{adilazuarda-etal-2022-indorobusta} and the specific needs of low-resource languages \cite{cahyawijaya-etal-2023-nusawrites}.

Furthermore, while efforts like XLM-R \cite{conneau-etal-2020-unsupervised} and mBERT \cite{devlin2018bert} have aimed to introduce cross-linguality, their focus on achieving state-of-the-art performance in cross-lingual language understanding tasks may not fully address the unique issues faced by language models operating within Indonesia’s complex multilingual and multicultural environment. \citet{cahyawijaya-etal-2023-nusawrites} show that even these large multilingual models fail to outperform classical baselines on extremely low-resource languages.

In light of this, the successful enhancement of a Thai language model through PhayaThaiBERT \cite{sriwirote2023phayathaibert} serves as a compelling source of inspiration for addressing these challenges. PhayaThaiBERT achieved significant advancements by specifically addressing the challenge of unassimilated loanwords. This approach is particularly relevant to the context of Indonesia, as low-resource regional languages are often influenced by other languages through code-switching, introducing a significant number of unassimilated loanwords into their vocabulary.

Motivated by the successful approach of PhayaThaiBERT and its potential for adaptation to low-resource languages, we propose NusaBERT. This novel model builds upon the foundation laid by IndoBERT \cite{wilie-etal-2020-indonlu} and targets the linguistic complexities of low-resource regional languages in Indonesia. NusaBERT also leverages techniques inspired by PhayaThaiBERT, such as vocabulary expansion, and aims to achieve state-of-the-art performance on various multilingual benchmark datasets.

\section{Related Works}

Recent years have witnessed significant progress in Indonesian NLP research. Pre-trained language models like IndoBERT \cite{wilie-etal-2020-indonlu} and IndoBART \cite{cahyawijaya-etal-2021-indonlg} have demonstrated the effectiveness of this approach for various Indonesian language tasks. IndoBERT, built on the principles of Google’s BERT architecture \cite{devlin2018bert}, was specifically trained on a large Indonesian text corpus. It achieved state-of-the-art performance on the IndoNLU benchmark \cite{wilie-etal-2020-indonlu}, a collection of Indonesian-specific NLU tasks like text classification, question answering, and named entity recognition, demonstrating its competence in understanding the nuances of the Indonesian language. IndoBART, based on the BART architecture \cite{lewis-etal-2020-bart}, focuses on sequence-to-sequence tasks within the Indonesian language. This model has found success in language generation tasks like machine translation and text summarization, highlighting its ability to process and produce natural Indonesian text.

NusaX \cite{winata-etal-2023-nusax}, a benchmark for 10 under-resourced Indonesian local languages, shows that when IndoBERT and IndoBART are fine-tuned for these languages, they achieve impressive results in sentiment analysis and machine translation, respectively. Specifically, $\text{IndoBERT}_{\text{LARGE}}$ performs comparably to the larger, multilingual $\text{XLM-R}_{\text{LARGE}}$ in sentiment analysis, despite the latter's exposure to some of these languages during pre-training. In machine translation tasks, IndoBART excels in translating from local languages to Indonesian, closely followed by PBSMT. These findings show the transferability of Indonesian with several regional languages since they share similar characteristics. It also demonstrates how a smaller monolingual model can be used to replace large multilingual models for local-language tasks in the Indonesian linguistic landscape in certain cases.

Afterward, NusaWrites \cite{cahyawijaya-etal-2023-nusawrites} was released and complements NusaX by providing a more lexically diverse and culturally relevant dataset on 12 underrepresented local languages. The study compared various data collection methods such as online scraping from Wikipedia, human translation, and paragraph writing by native speakers. It shows that paragraph writing is the most promising strategy to build a high-quality and culturally relevant corpus. The paper presents a new high-quality benchmark based on human-annotated texts produced from topic-focused paragraph writing (NusaParagraph) and human translation (NusaTranslation).

Upon fine-tuning different models on these new benchmarks, results show that multilingual models like XLM-R \cite{conneau-etal-2020-unsupervised} and mBERT \cite{devlin2018bert} and monolingual models (i.e., IndoBERT, IndoBART, and IndoGPT) fail to outperform classical machine learning models on several extremely low-resource languages. Aside from languages that were included in the pre-training corpus, only languages that are linguistically similar to Indonesian (\verb|ind|) such as Betawi (\verb|bew|) and Minangkabau (\verb|min|) benefit from the cross-lingual transfer.

Moreover, these models face limitations when addressing the full complexity of the Indonesian linguistic landscape, particularly concerning code-switching. \citet{zhang-etal-2023-multilingual} highlights a critical limitation: even multilingual pre-trained language models struggle with this prevalent phenomenon in Indonesian day-to-day communication, where speakers seamlessly interweave multiple languages within a single utterance. This suggests that the specific challenges of code-switching require models trained and tailored specifically for this phenomenon, even within the context of multilingual models. The limitations of existing models are further emphasized by studies like IndoRobusta \cite{adilazuarda-etal-2022-indorobusta}, which showcase their shortcomings in handling code-switching tasks effectively.

The success of PhayaThaiBERT \cite{sriwirote2023phayathaibert}, a Thai language model specifically designed to address the challenge of unassimilated loanwords, offers valuable inspiration for tackling challenges faced by NLP models in Indonesia. Similar to Thai, low-resource regional languages in Indonesia are frequently influenced by other languages due to code-switching. This phenomenon leads to a significant number of unassimilated loanwords, which are words from other languages adopted into the regional language but fully integrated into its grammar. These unassimilated loanwords pose a specific challenge for NLP models, as they often lack the necessary training data and knowledge to understand their context accurately. PhayaThaiBERT addresses this challenge by incorporating techniques such as vocabulary expansion. This technique involves augmenting the model’s vocabulary with these loanwords and variations, allowing it to better recognize and understand them within the context of the regional language.

Building upon the concept of vocabulary expansion, IndoBERTweet \cite{koto-etal-2021-indobertweet}, an extension of IndoLEM's IndoBERT \cite{koto-etal-2020-indolem}, tackles the challenge of informal language and social media slang by specifically augmenting its vocabulary with terms commonly found in Indonesian Twitter data. This vocabulary expansion helps the model better understand and process the nuances of informal communication, which often deviates from standard Indonesian grammar and incorporates unique slang terms. By incorporating vocabulary specific to a particular domain or usage context, models like IndoBERTweet can improve their performance on tasks like sentiment analysis or text classification within that domain, while also achieving significantly faster pre-training compared to training from scratch.

\section{NusaBERT}

This section introduces the vocabulary expansion method applied to IndoBERT \cite{wilie-etal-2020-indonlu}, the corpus dataset used for training, and the continued pre-training procedure. Subsequently, we will evaluate our resultant models on downstream tasks to measure their natural language understanding, multilinguality, and multicultural capabilities.

\subsection{Vocabulary Expansion and Dataset}

\subsubsection{Pre-training Corpus}
\label{sec:corpus}

Following the approach proposed by PhayaThaiBERT \cite{sriwirote2023phayathaibert}, we expanded IndoBERT’s \cite{wilie-etal-2020-indonlu} vocabulary to introduce foreign tokens and conducted continued pre-training with the expanded vocab size and word embeddings.

Specifically, as we aim to extend IndoBERT’s multilingual capabilities to regional languages of Indonesia, we first gathered open-source corpora that contain monolingual texts in the many languages of Indonesia. For instance, Indo4B+ \cite{cahyawijaya-etal-2021-indonlg} covers Indonesian (\verb|ind|), Javanese (\verb|jav|), and Sundanese (\verb|sun|), although the majority of the corpus is still mainly Indonesian. It is also mostly just CC-100 \cite{wenzek-etal-2020-ccnet} and Wikipedia subsets for Javanese and Sundanese. Much larger corpora like mC4 \cite{raffel2023exploring} and OSCAR \cite{suarez2019asynchronous} similarly only support Indonesian, Javanese, and Sundanese. The quality of these large-scale crawled corpora also requires further inspection and improvement \cite{kreutzer-etal-2022-quality}. More recent corpus like MADLAD-400 \cite{kudugunta2024madlad} support an even wider range of languages and while some level of data filters have been applied to help clean the dataset, the corpus is still naturally noisy.

For these reasons and to ensure the quality of our pre-training corpus, we opted for a much cleaner and stricter corpus like CulturaX \cite{nguyen2023culturax} which utilizes mC4 \cite{raffel2023exploring} and OSCAR \cite{suarez2019asynchronous} and applies rigorous filters. Through those filters and deduplication methods, CulturaX resolves the quality and duplication issues of naively merging raw mC4 and OSCAR. However, in terms of language coverage of languages of Indonesia, CulturaX only contains Indonesian (\verb|ind|), Javanese (\verb|jav|), and Sundanese (\verb|sun|). Further, we also decided to include Standard Malay (\verb|msa|) as the language is still spoken especially on the eastern coast of Sumatra and West Kalimantan \cite{wahyudibahasa,corporation2007world}.

To compensate for the lack of language diversity, we decided to leverage Wikipedia texts as the web contains even more regional languages of Indonesia. We used an open-source Wikipedia dataset\footnote{\url{https://huggingface.co/datasets/sabilmakbar/indo_wiki}} that is tailored specifically for the languages of Indonesia. The author of the dataset conveniently provides cleaned versions of the corpus which has been deduplicated and standardized into readable texts. The list of languages whose Wikipedia dumps are found in this dataset includes Acehnese (\verb|ace|), Balinese (\verb|ban|), Banjarese (\verb|bjn|), Banyumasan (\verb|jav|\footnote{A dialect of Javanese (\texttt{jav}), sometimes assigned the ISO code \texttt{map-bms}.}), Buginese (\verb|bug|), Gorontalo (\verb|gor|), Indonesian (\verb|ind|), Javanese (\verb|jav|), Minangkabau (\verb|min|), Malay (\verb|msa|), Nias (\verb|nia|), Sundanese (\verb|sun|), and Tetum (\verb|tet|). Like Malay (\verb|msa|), Tetum (\verb|tet|) was included in our study as it remains a minority regional language in West Timor, Indonesia. 

Finally, since the Wikipedia dataset is rather small in quantity compared to that of typical common crawl corpora, we decided to also include subsets of the NLLB corpus \cite{costa2022no} that similarly employ rigorous data filtering methods. Likewise, we used an open-source modification of the NLLB corpus called KoPI-NLLB\footnote{\url{https://huggingface.co/datasets/acul3/KoPI-NLLB}} that conveniently selects subsets of languages of Indonesia and applies additional deduplication procedures. KoPI-NLLB has subsets for Acehnese (\verb|ace|), Balinese (\verb|ban|), Banjarese (\verb|bjn|), Indonesian (\verb|ind|), Javanese (\verb|jav|), Minangkabau (\verb|min|), and Sundanese (\verb|sun|). However, since CulturaX \cite{nguyen2023culturax} already has a significant percentage of Indonesian texts, we do not include the Indonesian subset of KoPI-NLLB and retain only the regional language subsets.

Our pre-training corpus thereby consists of 13 languages and is a combination of CulturaX \cite{nguyen2023culturax}, Wikipedia, and KoPI-NLLB, with a strong emphasis on data quality via strict filtering and deduplication methods. A summary of the overall dataset and list of languages included in the pre-training corpus can be found in Table \ref{corpus-language-summary}.

\begin{table}[t]
\centering
\def\arraystretch{1.2}
\begin{tabular}{lr}
\hline
\textbf{Language (ISO 639-3)} & \textbf{\#documents} \\
\hline
Indonesian (\verb|ind|)      & 23,905,655           \\
Javanese (\verb|jav|)        & 1,229,867            \\
Sundanese (\verb|sun|)       & 957,674              \\
Acehnese (\verb|ace|)        & 805,498              \\
Malay (\verb|msa|)           & 584,186              \\
Minangkabau (\verb|min|)     & 339,181              \\
Banjarese (\verb|bjn|)       & 306,751              \\
Balinese (\verb|ban|)        & 264,382              \\
Gorontalo (\verb|gor|)       & 14,514               \\
Banyumasan (\verb|jav|)  & 11,832               \\
Buginese (\verb|bug|)        & 9,793                \\
Nias (\verb|nia|)            & 1,650                \\
Tetum (\verb|tet|)           & 1,465                \\
\hline
Total       & 28,432,448              \\
\hline
\end{tabular}
\caption{\label{corpus-language-summary} A summary of the number of documents per language in the pre-training corpus of NusaBERT.}
\end{table}

\subsubsection{Vocabulary Expansion}
\label{sec:tokenizer}

Unlike PhayaThaiBERT \cite{sriwirote2023phayathaibert}, we did not transfer the non-overlapping vocabulary of XLM-R \cite{conneau-etal-2020-unsupervised}. Instead, we decided to train a new WordPiece tokenizer \cite{wu2016googles} based on the IndoBERT tokenizer \cite{wilie-etal-2020-indonlu} on the newly formed corpus. There are several design choices considered when training the new tokenizer, such as the target vocabulary size and the subsets to be included during tokenizer training.

For the latter, we decided not to include the Indonesian subset of CulturaX due to its large percentage and that it would diminish the importance of non-Indonesian tokens, which contradicts the main goal of NusaBERT. However, the relatively smaller Indonesian Wikipedia is still included as there might be newer Indonesian words and terms that might have not been included in the current IndoBERT tokenizer.

On the other hand, for the former, we followed a close estimate to that of Typhoon language models \cite{pipatanakul2023typhoon} whose design choice is based on another previous study that investigated the most efficient target vocabulary size \cite{csaki2023efficiently}. Both studies suggested a vocabulary size of 5,000, but our preliminary experiments found that a target vocabulary size of 5,000 has very few new tokens to be added to the current tokenizer. Due to this, we increased the target vocabulary size to 10,000 and found 1,511 new, non-overlapping tokens to be added.

While this increase is not as significant as originally proposed in PhayaThaiBERT \cite{sriwirote2023phayathaibert}, we considered the downstream effects of significantly increasing the number of parameters if we decided to exactly follow their approach. Moreover, WangchanBERTa \cite{lowphansirikul2021wangchanberta}, the base model of PhayaThaiBERT, has a deeper issue of only supporting mainly Thai tokens and struggles with unassimilated loanwords in the Latin alphabet. The IndoBERT tokenizer, on the other hand, has been trained on an Indonesian corpus that uses the Latin alphabet and NusaBERT aims to only introduce regional language tokens. Therefore, we finalized this set of additional tokens which increased IndoBERT’s vocabulary size from 30,521 to 32,032.

\subsection{Continued Pre-training}
\label{sec:continued-pre-training}

\subsubsection{Model Configuration and Initialization}

Like PhayaThaiBERT \cite{sriwirote2023phayathaibert}, we conducted continued pre-training with the initial model checkpoints of IndoBERT \cite{wilie-etal-2020-indonlu}. We experimented with two size variants of IndoBERT, namely $\text{IndoBERT}_{\text{BASE}}$ and $\text{IndoBERT}_{\text{LARGE}}$. In both variants, we used phase one checkpoints of IndoBERT instead of phase two checkpoints. Therefore, the initial parameters of our model are identical to that of IndoBERT with the exception of the new vocabulary’s embeddings, which are initialized from the mean of the old word embeddings \cite{hewitt2021initializing}. There are no additional architectural changes added to the original BERT architecture \cite{devlin2018bert} and we thereby call our new extended models $\text{NusaBERT}_{\text{BASE}}$ (111M parameters) and $\text{NusaBERT}_{\text{LARGE}}$ (337M parameters), respectively.

\subsubsection{Data Pre-processing}

During the continued pre-training, we decided to keep the same sequence length of 128 as IndoBERT phase one models. Our data pre-processing procedures follow a typical masked language modeling pre-processing setup. Firstly, a random 5\% sample of our corpus described in Section \S \ref{sec:corpus} is held out for evaluation purposes. Secondly, all texts are tokenized using the newly extended tokenizer described in \S \ref{sec:tokenizer}. Since our tokenizer follows exactly from the original IndoBERT tokenizer, special tokens such as \verb|[CLS]| and \verb|[SEP]| are added at the start and end of all texts. Finally, batches of tokenized texts are then concatenated into one long sequence and then grouped into sequences of length 128 tokens each. Sequences shorter than 128 are thus discarded. These batches of fixed-length tokenized sequences are thereby ready for training purposes.

\subsubsection{Pre-training Objective and Procedures}

Instead of using the original BERT \cite{devlin2018bert} objective of both next sentence prediction (NSP) and masked language modeling (MLM), we opted for the more robust RoBERTa \cite{liu2019roberta} objective that drops the NSP objective and retains only the MLM objective. We used the same masking sampling technique employed in RoBERTa, whereby 15\% of tokens in each sequence are either replaced with a special \verb|[MASK]| token (0.8 probability), replaced with another random token (0.1 probability), or kept as is (0.1 probability). Unlike BERT, the RoBERTa-style masking is applied dynamically instead of statically for every batch of text during pre-training.

With this setup, we conducted continued pre-training for 500,000 optimization steps with hyperparameters shown in Table \ref{pretraining-hyperparam}. Unlike PhayaThaiBERT \cite{sriwirote2023phayathaibert}, our continued pre-training procedure doesn’t involve sophisticated fine-tuning techniques. Instead, we simply trained our models with 24,000 warmup steps to the peak learning rate and applied a linear learning rate decay to zero, with a batch size of 256 on a single GPU. Effectively, our model was trained on 16B tokens.

\begin{table}[t]
\centering
\def\arraystretch{1.2}
\begin{tabular}{lc}
\hline
\textbf{Hyperparameter} & \textbf{Value} \\
\hline
Sequence length         & 128                      \\
Batch size              & 256                      \\
Peak learning rate      & {[}3e-4, 3e-5{]}\textsuperscript{*}        \\
\#warmup steps          & 24,000                   \\
\#optimization steps    & 500,000                  \\
Learning rate scheduler & Linear                   \\
Optimizer               & AdamW                    \\
Adam $(\beta_1, \beta_2)$ & (0.9, 0.999)           \\
Adam $\epsilon$         & 1e-8                     \\
Weight decay            & 0.01                     \\
PyTorch data type       & bfloat16                 \\
\hline
\end{tabular}
\caption{\label{pretraining-hyperparam} Continued pre-training hyperparameters. \\ \textsuperscript{*} indicates the differing values for $\text{NusaBERT}_{\text{BASE}}$ and $\text{NusaBERT}_{\text{LARGE}}$, respectively.}
\end{table}

We hypothesize that our training procedure is stable, unlike PhayaThaiBERT, due to the latter’s massive increase in vocabulary size which may have caused training instabilities. Conversely, as NusaBERT only introduces 1,511 new tokens, our model training is relatively stable and simpler.

\subsection{Evaluation Benchmark}
\label{sec:evaluation-benchmark}

Our benchmark concentrates on three aspects: (1) natural language understanding (NLU), (2) multilinguality, and (3) multicultural. Therefore, we decided to utilize the Indonesian NLU benchmark IndoNLU \cite{wilie-etal-2020-indonlu}, and multilingual NLU benchmarks such as NusaX \cite{winata-etal-2023-nusax}, and NusaWrites \cite{cahyawijaya-etal-2023-nusawrites} which contain a wide range of regional languages of Indonesia and closely reflect the local cultures.

\begin{table}[t]
\small
\centering
\def\arraystretch{1.2}
\begin{tabular}{ll}
\hline
\textbf{Dataset} & \textbf{Task} \\
\hline
\\[-1.2em]
\rowcolor{ivory2}
\multicolumn{2}{c}{\textit{\textbf{Single-sentence Classification}}} \\
EmoT & Emotion Classification \\
SmSA & Sentiment Analysis \\
NusaX  & Sentiment Analysis \\
NusaT Sentiment & Sentiment Analysis \\
NusaT Emotion & Emotion Classification \\
NusaP Emotion & Emotion Classification \\
NusaP Rhetorical & Rhetorical Mode Classification \\
NusaP Topic & Topic Modeling \\
\hline
\\[-1.2em]
\rowcolor{ivory2}
\multicolumn{2}{c}{\textit{\textbf{Single-sentence Multi-label Classification}}} \\
CASA & Aspect-based Sentiment Analysis \\
HoASA & Aspect-based Sentiment Analysis \\
\hline
\\[-1.2em]
\rowcolor{ivory2}
\multicolumn{2}{c}{\textit{\textbf{Sequence-pair Classification}}} \\
WReTE & Textual Entailment \\
\hline
\\[-1.2em]
\rowcolor{ivory2}
\multicolumn{2}{c}{\textit{\textbf{Token Classification}}} \\
POSP & Part-of-Speech Tagging \\
BaPOS & Part-of-Speech Tagging \\
TermA & Span Extraction \\
KEPS & Span Extraction \\
NERGrit & Named Entity Recognition \\
NERP & Named Entity Recognition \\
\hline
\\[-1.2em]
\rowcolor{ivory2}
\multicolumn{2}{c}{\textit{\textbf{Sequence-Pair Token Classification}}} \\
FacQA & Span Extraction \\
\hline
\end{tabular}
\caption{\label{evaluation-benchmarks} List of downstream evaluation benchmarks for NusaBERT fine-tuning.}
\end{table}

In general, the tasks in these benchmarks can be divided into five major categories: (a) single-sentence classification, (b) single-sentence multi-label classification, (c) sequence-pair classification, (d) token classification, and (e) sequence-pair token classification.

\subsubsection{Datasets}

The IndoNLU benchmark \cite{wilie-etal-2020-indonlu} consists only of Indonesian datasets from various NLU tasks. On the other hand, NusaX \cite{winata-etal-2023-nusax} and NusaWrites \cite{cahyawijaya-etal-2023-nusawrites} provide NLU benchmarks for a variety of regional languages of Indonesia like Ambon, Toba Batak, Betawi, etc.  A high-level overview of the benchmarks is shown in Table \ref{evaluation-benchmarks}. The list of all languages and dialects involved in this study and its details are found in Appendix \ref{sec:language-studied}.

\paragraph{IndoNLU}
IndoNLU \cite{wilie-etal-2020-indonlu} is a comprehensive benchmark corpus designed to facilitate research in Indonesian natural language understanding. It comprises multiple datasets covering a variety of NLU tasks, which can be categorized into two main tasks: text classification and sequence labeling. The benchmark aims to provide a standard for evaluating the performance of models on Indonesian language tasks, addressing the need for more resources in languages other than English. The dataset supports text classification tasks like emotion classification, sentiment analysis, textual entailment, and aspect-based sentiment analysis (ABSA) making it versatile for testing different aspects of language understanding models. Further, the sequence labeling datasets include subtasks such as part-of-speech tagging, span extraction, and named entity recognition.

\paragraph{NusaX}
NusaX \cite{winata-etal-2023-nusax} is a multilingual benchmark that focuses on assessing the capabilities of NLU performance of language models across 10 low-resource local Indonesian languages, with the addition of Indonesian and English. The dataset was originally the IndoNLU’s SmSA sentiment analysis dataset \cite{wilie-etal-2020-indonlu}, which was then translated into 11 other languages. Its main task is therefore sentiment analysis, although the dataset is likewise usable for machine translation purposes. For the evaluation of our model, we utilized the sentiment analysis dataset only.

\paragraph{NusaWrites}
NusaWrites \cite{cahyawijaya-etal-2023-nusawrites} is a multilingual benchmark that serves as an extension of NusaX \cite{winata-etal-2023-nusax} and encompasses 12 underrepresented and low-resource languages in Indonesia. By its design, NusaWrites is more locally nuanced than generic corpora like Wikipedia and is lexically more diverse. It contains 2 sub-corpus defined by the way the data is constructed, topic-focused paragraph writing from human annotators (NusaParagraph) and human translation by native speakers (NusaTranslation). NusaParagraph contains three downstream tasks which include topic classification, emotion classification, and rhetoric mode classification. On the other hand, NusaTranslation contains three parallel downstream tasks which are sentiment analysis, emotion classification, and machine translation. Like NusaX, NusaTranslation is a human-translated version of IndoNLU’s EmoT emotion classification dataset \cite{wilie-etal-2020-indonlu} and IndoLEM’s sentiment analysis dataset \cite{koto-etal-2020-indolem}. Again, we leave out the machine translation dataset from NusaTranslation as it is irrelevant to our study.

\subsubsection{Benchmarking Models}

\begin{table*}[t]
\centering
\def\arraystretch{1.2}
\begin{tabular}{lccc}
\hline
\textbf{Hyperparameter} & \textbf{Sentence Classification} & \textbf{Multi-label Classification} & \textbf{Token Classification} \\
\hline
\#epochs & 100 & 100 & 10 \\
Learning rate & {[}1e-5, 2e-5{]}\textsuperscript{*} & 1e-5 & 2e-5 \\
Weight decay & 0.01 & 0.01 & 0.01 \\
Batch size & {[}32, 16{]}\textsuperscript{*} & 32 & 16 \\
\hline
\end{tabular}
\caption{\label{fine-tuning-hyperparam} Downstream fine-tuning hyperparameters. \textsuperscript{*} indicates the differing values for $\text{NusaBERT}_{\text{BASE}}$ and $\text{NusaBERT}_{\text{LARGE}}$, respectively.}
\end{table*}

We compared the performance of our NusaBERT models against the official report benchmark results without any further fine-tuning of the baseline models. The IndoNLU benchmark results \cite{wilie-etal-2020-indonlu} include monolingual Indonesian language models $\text{IndoBERT}_{\text{BASE}}$, $\text{IndoBERT}_{\text{LARGE}}$, $\text{IndoBERT-lite}_{\text{BASE}}$, $\text{IndoBERT-lite}_{\text{LARGE}}$, as well as multilingual language models like mBERT \cite{devlin2018bert}, XLM-MLM \cite{conneau2019cross}, and both $\text{XLM-R}_{\text{BASE}}$ and $\text{XLM-R}_{\text{LARGE}}$ \cite{conneau-etal-2020-unsupervised}. Additionally, NusaX \cite{winata-etal-2023-nusax} and NusaWrites \cite{cahyawijaya-etal-2023-nusawrites} report on the same set of models, with the inclusion of the IndoLEM IndoBERT \cite{koto-etal-2020-indolem}, and traditional/classical machine learning models such as Naive Bayes, SVM, and Logistic Regression.

\subsubsection{Fine-Tuning Setup}

To fairly compare our results with the baselines, we adhere to similar fine-tuning procedures outlined in their respective benchmark codebases. Table \ref{fine-tuning-hyperparam} details the hyperparameters employed for fine-tuning the models across various tasks, reflecting the benchmarks' recommended settings with minor adjustments to learning rates and batch sizes for certain tasks.

For IndoNLU, NusaX, and NusaTranslation benchmarks, we used a sequence length of 128, while for NusaParagraph, we increased the sequence length to 512 due to its much longer input text length. We applied early stopping based on the evaluation metrics and chose the best-scoring model. All fine-tuning processes utilize the Trainer API from Hugging Face’s transformers library \cite{wolf-etal-2020-transformers}. For other hyperparameters not mentioned in Table \ref{fine-tuning-hyperparam}, we followed the default hyperparameter from the Trainer API.

\subsubsection{Evaluation Metrics}

We evaluated the performance of our fine-tuned models using the macro-averaged F1 score for classification tasks, as specified in the IndoNLU, NusaX, and NusaWrites. Likewise, we followed the sequence labeling evaluation procedure used for CoNLL for token classification tasks of IndoNLU.

\section{Results and Analysis}

\subsection{Pre-training Results}

Both $\text{NusaBERT}_{\text{BASE}}$ and $\text{NusaBERT}_{\text{LARGE}}$ converged smoothly during the continued pre-training phase detailed in \S \ref{sec:continued-pre-training}. Particularly, after 500,000 steps, $\text{NusaBERT}_{\text{BASE}}$ achieved an evaluation loss of 1.488 (4.427 PPL). Similarly, $\text{NusaBERT}_{\text{LARGE}}$ achieved a lower evaluation loss of 1.327 (3.769 PPL).

\subsection{Fine-tuning Results}

As outlined in \S \ref{sec:evaluation-benchmark}, we fine-tuned NusaBERT models on several downstream tasks in Indonesian and regional languages. Since each benchmark contains several subsets, we will thoroughly discuss our model’s results on each benchmark in the following sections.

\paragraph{IndoNLU}

\begin{table*}[t]
\scriptsize
\centering
\setlength{\tabcolsep}{0.6\tabcolsep}
\def\arraystretch{1.2}
\begin{tabular}{lcccccc|cccccccc}
\hline
\multirow{2}{*}{\textbf{Model}} & \multicolumn{6}{c}{\textbf{Classification}} & \multicolumn{8}{c}{\textbf{Sequence Labeling}} \\ \cline{2-15} 
 & \textbf{EmoT} & \textbf{SmSA} & \textbf{CASA} & \textbf{HoASA} & \textbf{WReTE} & $\mu$ & \textbf{POSP} & \textbf{BaPOS} & \textbf{TermA} & \textbf{KEPS} & \textbf{NERGrit} & \textbf{NERP} & \textbf{FacQA} & $\mu$ \\ \hline
mBERT & 67.30 & 84.14 & 72.23 & 84.63 & 84.40 & 78.54 & 91.85 & 83.25 & 89.51 & 64.31 & 75.02 & 69.27 & 61.29 & 76.36 \\
XLM-MLM & 65.75 & 86.33 & 82.17 & 88.89 & 64.35 & 77.50 & 95.87 & 88.40 & 90.55 & 65.35 & 74.75 & 75.06 & 62.15 & 78.88 \\
$\text{XLM-R}_{\text{BASE}}$ & 71.15 & 91.39 & 91.71 & 91.57 & 79.95 & 85.15 & 95.16 & 84.64 & 90.99 & 68.82 & 79.09 & 75.03 & 64.58 & 79.76 \\
$\text{XLM-R}_{\text{LARGE}}$ & 78.51 & 92.35 & 92.40 & \textbf{94.27} & 83.82 & 88.27 & 92.73 & 87.03 & 91.45 & 70.88 & 78.26 & 78.52 & \textbf{74.61} & 81.93 \\
\hline
$\text{IndoBERT-lite}_{\text{BASE}}$ & 73.88 & 90.85 & 89.68 & 88.07 & 82.17 & 84.93 & 91.40 & 75.10 & 89.29 & 69.02 & 66.62 & 46.58 & 54.99 & 70.43 \\
\quad+ phase two & 72.27 & 90.29 & 87.63 & 87.62 & 83.62 & 84.29 & 90.05 & 77.59 & 89.19 & 69.13 & 66.71 & 50.52 & 49.18 & 70.34 \\
$\text{IndoBERT-lite}_{\text{LARGE}}$ & 75.19 & 88.66 & 90.99 & 89.53 & 78.98 & 84.67 & 91.56 & 83.74 & 90.23 & 67.89 & 71.19 & 74.37 & 65.50 & 77.78 \\
\quad+ phase two & 70.80 & 88.61 & 88.13 & 91.05 & \textbf{85.41} & 84.80 & 94.53 & 84.91 & 90.72 & 68.55 & 73.07 & 74.89 & 62.87 & 78.51 \\
$\text{IndoBERT}_{\text{BASE}}$ & 75.48 & 87.73 & 93.23 & 92.07 & 78.55 & 85.41 & 95.26 & 87.09 & 90.73 & 70.36 & 69.87 & 75.52 & 53.45 & 77.47 \\
\quad+ phase two & 76.28 & 87.66 & 93.24 & 92.70 & 78.68 & 85.71 & 95.23 & 85.72 & 91.13 & 69.17 & 67.42 & 75.68 & 57.06 & 77.34 \\
$\text{IndoBERT}_{\text{LARGE}}$ & 77.08 & \textbf{92.72} & \textbf{95.69} & 93.75 & 82.91 & \textbf{88.43} & 95.71 & 90.35 & 91.87 & 71.18 & 77.60 & 79.25 & 62.48 & 81.21 \\
\quad+ phase two & \textbf{79.47} & 92.03 & 94.94 & 93.38 & 80.30 & 88.02 & 95.34 & 87.36 & \textbf{92.14} & 71.27 & 76.63 & 77.99 & 68.09 & 81.26 \\
\hline
$\text{NusaBERT}_{\text{BASE}}$ & 76.10 & 87.46 & 91.26 & 89.80 & 76.77 & 84.28 & 95.77 & 96.02 & 90.54 & 66.67 & 72.93 & 82.29 & 54.81 & 79.86 \\
$\text{NusaBERT}_{\text{LARGE}}$ & 78.90 & 87.36 & 92.13 & 93.18 & 82.64 & 86.84 & \textbf{96.89} & \textbf{96.76} & 91.73 & \textbf{71.53} & \textbf{79.86} & \textbf{85.12} & 66.77 & \textbf{84.09} \\
\hline
\end{tabular}
\caption{\label{indonlu-results} Evaluation results of baseline models and NusaBERT on the IndoNLU benchmark tasks, measured in macro-F1 (\%). Baseline results are obtained from \citet{wilie-etal-2020-indonlu}. The best performance on each task is \textbf{bolded} for clarity.}
\end{table*}

\begin{table*}[t]
\small
\centering
\setlength{\tabcolsep}{0.9\tabcolsep}
\def\arraystretch{1.2}
\begin{tabular}{lccccccccccccc}
\hline
\textbf{Model} & \textbf{ace} & \textbf{ban} & \textbf{bbc} & \textbf{bjn} & \textbf{bug} & \textbf{eng} & \textbf{ind} & \textbf{jav} & \textbf{mad} & \textbf{min} & \textbf{nij} & \textbf{sun} & $\mu$ \\ \hline
Logistic Regression & 77.4 & 76.3 & 76.3 & 75.0 & 77.2 & 75.9 & 74.7 & 73.7 & 74.7 & 74.8 & 73.4 & 75.8 & 75.4 \\
Naive Bayes & 72.5 & 72.6 & 73.0 & 71.9 & 73.7 & 76.5 & 73.1 & 69.4 & 66.8 & 73.2 & 68.8 & 71.9 & 72.0 \\
SVM & 75.7 & 75.3 & \textbf{76.7} & 74.8 & \textbf{77.2} & 75.0 & 78.7 & 71.3 & 73.8 & 76.7 & 75.1 & 74.3 & 75.4 \\ \hline
mBERT & 72.2 & 70.6 & 69.3 & 70.4 & 68.0 & 84.1 & 78.0 & 73.2 & 67.4 & 74.9 & 70.2 & 74.5 & 72.7 \\
$\text{XLM-R}_{\text{BASE}}$ & 73.9 & 72.8 & 62.3 & 76.6 & 66.6 & 90.8 & 88.4 & 78.9 & 69.7 & 79.1 & 75.0 & 80.1 & 76.2 \\
$\text{XLM-R}_{\text{LARGE}}$ & 75.9 & 77.1 & 65.5 & 86.3 & 70.0 & \textbf{92.6} & 91.6 & 84.2 & 74.9 & 83.1 & 73.3 & \textbf{86.0} & 80.0 \\ \hline
$\text{IndoLEM IndoBERT}_{\text{BASE}}$ & 72.6 & 65.4 & 61.7 & 71.2 & 66.9 & 71.2 & 87.6 & 74.5 & 71.8 & 68.9 & 69.3 & 71.7 & 71.1 \\
$\text{IndoNLU IndoBERT}_{\text{BASE}}$ & 75.4 & 74.8 & 70.0 & 83.1 & 73.9 & 79.5 & 90.0 & 81.7 & 77.8 & 82.5 & 75.8 & 77.5 & 78.5 \\
$\text{IndoNLU IndoBERT}_{\text{LARGE}}$ & 76.3 & 79.5 & 74.0 & 83.2 & 70.9 & 87.3 & 90.2 & 85.6 & 77.2 & 82.9 & 75.8 & 77.2 & 80.0 \\ \hline
$\text{NusaBERT}_{\text{BASE}}$ & 76.5 & 78.7 & 74.0 & 82.4 & 71.6 & 84.1 & 89.7 & 84.1 & 75.6 & 80.8 & 74.9 & 85.2 & 79.8 \\
$\text{NusaBERT}_{\text{LARGE}}$ & \textbf{81.8} & \textbf{82.8} & 74.7 & \textbf{86.5} & 73.4 & 84.6 & \textbf{93.3} & \textbf{87.2} & \textbf{82.5} & \textbf{83.5} & \textbf{77.7} & 82.7 & \textbf{82.6} \\
\hline
\end{tabular}
\caption{\label{nusax-results} Evaluation results of baseline models and NusaBERT on the NusaX sentiment analysis task, measured in macro-F1 (\%). Baseline results are obtained from \citet{winata-etal-2023-nusax}. The best performance on each task is \textbf{bolded} for clarity.}
\end{table*}

We report the official baseline results as well as the results of NusaBERT in Table \ref{indonlu-results}. As shown, our models’ performance on classification tasks of IndoNLU slightly deteriorates from that of the original IndoBERT models. The overall average score of NusaBERT decreases by about 1-2\%, with $\text{NusaBERT}_{\text{BASE}}$ decreasing from 85.41\% to 84.28\% (\minus1.13\%) and $\text{NusaBERT}_{\text{LARGE}}$ decreasing from 88.43\% to 86.84\% (\minus1.59\%). Our models struggle particularly with aspect-based sentiment analysis tasks (CASA and HoASA), and the $\text{NusaBERT}_{\text{LARGE}}$ result on SmSA drops by about 5\%.

In contrast, NusaBERT significantly improves the sequence labeling results of IndoBERT, increasing the average score by about 2-3\%. $\text{NusaBERT}_{\text{BASE}}$ improves the base IndoBERT model score from 77.47\% to 79.86\% (\plus2.39\%), while $\text{NusaBERT}_{\text{LARGE}}$ improves the score from 81.21\% to 84.09\% (\plus2.88\%). NusaBERT especially improves the results on part-of-speech tagging tasks (POSP, BaPOS) and named entity recognition tasks (NERGrit, NERP).

Further, since the results of IndoBERT are similar to those of multilingual models like XLM-R, we observed a similar trend when comparing NusaBERT with the latter. That is, our models are slightly worse on classification tasks (\minus0.87\% $\text{NusaBERT}_{\text{BASE}}$, \minus1.43\% $\text{NusaBERT}_{\text{LARGE}}$), yet better on sequence labeling tasks (\plus0.1\% $\text{NusaBERT}_{\text{BASE}}$, \plus2.17\% $\text{NusaBERT}_{\text{LARGE}}$) when compared against XLM-R.

These indicate that our models remain competitive on Indonesian NLU tasks, retaining most of its initial knowledge found in the base IndoBERT model. Further experiments are required to fully retain and improve the results of IndoBERT across all tasks while still introducing multilingual capabilities to NusaBERT.

\paragraph{NusaX}

\begin{table*}[t]
\centering
\def\arraystretch{1.2}
\begin{tabular}{lccc|cc}
\hline
\multirow{2}{*}{\textbf{Model}} & \multicolumn{3}{c}{\textbf{NusaParagraph}} & \multicolumn{2}{c}{\textbf{NusaTranslation}} \\ \cline{2-6} 
 & \textbf{Emotion} & \textbf{Rhetorical Mode} & \textbf{Topic} & \textbf{Emotion} & \textbf{Sentiment} \\
 \hline
Logistic Regression & \textbf{78.23} & 45.21 & \textbf{87.67} & 56.18 & 74.89 \\
Naive Bayes & 75.51 & 37.73 & 85.06 & 52.70 & 74.89 \\
SVM & 76.36 & 45.44 & 85.86 & 55.08 & 76.04 \\ \hline
mBERT & 63.15 & 50.01 & 73.82 & 44.13 & 68.72 \\
$\text{XLM-R}_{\text{BASE}}$ & 59.15 & 49.17 & 71.68 & 47.02 & 68.62 \\
$\text{XLM-R}_{\text{LARGE}}$ & 67.42 & 51.57 & 83.05 & 54.84 & 79.06 \\  \hline
$\text{IndoLEM IndoBERT}_{\text{BASE}}$ & 66.94 & 51.93 & 84.87 & 52.59 & 69.08 \\
$\text{IndoNLU IndoBERT}_{\text{BASE}}$ & 67.12 & 47.92 & 85.87 & 54.50 & 75.24 \\
$\text{IndoNLU IndoBERT}_{\text{LARGE}}$ & 62.65 & 31.75 & 85.41 & 57.80 & 77.40 \\
$\text{NusaBERT}_{\text{BASE}}$ & 67.18 & 51.34 & 83.32 & 56.54 & 77.07 \\
$\text{NusaBERT}_{\text{LARGE}}$ & 71.82 & \textbf{53.06} & 85.08 & \textbf{61.40} & \textbf{79.54} \\
 \hline
\end{tabular}
\caption{\label{nusawrites-results} Evaluation results of baseline models and NusaBERT on the NusaWrites benchmark tasks, measured in macro-F1 (\%) and averaged over all of the languages found in each task. Detailed per-task and per-language results are shown in Appendix \ref{sec:nusawrites-complete-results}. Baseline results are obtained from \citet{cahyawijaya-etal-2023-nusawrites}. The best performance on each task is \textbf{bolded} for clarity.}
\end{table*}

The official baseline and NusaBERT results on NusaX are shown in Table \ref{nusax-results}. From the baseline result, the monolingual IndoBERT models outperformed larger multilingual models like mBERT and are on par against XLM-R models despite being trained only on Indonesian texts, suggesting strong transferability from Indonesian to regional languages \cite{winata-etal-2023-nusax}. It thus remains whether NusaBERT’s introduction to regional languages will benefit the model when fine-tuned on downstream, multilingual, regional language tasks.

On average, our models improve the results of both size-variants of IndoBERT. The $\text{NusaBERT}_{\text{BASE}}$ model increases the average score from 78.5\% to 79.8\% (\plus1.3\%) while $\text{NusaBERT}_{\text{LARGE}}$ increases the average score from 80.0\% to 82.6\% (\plus2.6\%). In particular, NusaBERT significantly improves the results on most languages that were included during the continued pre-training phase such as Acehnese (\verb|ace|), Balinese (\verb|ban|), Banjarese (\verb|bjn|), Buginese (\verb|bug|), Javanese (\verb|jav|), and Sundanese (\verb|sun|). However, this improvement is not consistent across all cases, particularly noting a slight decline in the performance of $\text{NusaBERT}_{\text{BASE}}$, even for languages included in the continued pre-training phase. Moreover, the results of languages not included in the continued pre-training phase like Madurese (\verb|mad|) and Ngaju (\verb|nij|) are still improved especially in $\text{NusaBERT}_{\text{LARGE}}$.

Overall, $\text{NusaBERT}_{\text{LARGE}}$ attained state-of-the-art results on most languages of NusaX, except for English (\verb|eng|) and Sundanese (\verb|sun|). XLM-R, which was pre-trained on these two languages \cite{conneau-etal-2020-unsupervised}, is unsurprisingly still best. Likewise, classical machine learning algorithms like SVM and Logistic Regression achieved the highest scores on Buginese (\verb|bug|) and Toba Batak (\verb|bbc|), two extremely low-resource languages. Our findings align with the suggestion of \citet{winata-etal-2023-nusax} whereby these languages are highly distinct from other languages of Indonesia and hence do not exhibit strong cross-lingual transferability. 

We also note that both languages stem from different language families than most of the other languages, even though they are all grouped into one Malayo-Polynesian subgroup \cite{ethnologue}. Buginese (\verb|bug|) is spoken mostly in the South Sulawesi region, while Toba Batak (\verb|bbc|) is spoken primarily in the Northwestern Sumatra and Barrier Islands regions.

In addition, while Buginese (\verb|bug|) is included in our pre-training corpus, it is the third smallest subset within our Wikipedia dataset, with only about 9,000 documents. Therefore, it remains our interest to find other ways to improve the results of languages that are not only extremely low-resource but are also highly distinct from other languages of Indonesia.

\paragraph{NusaWrites}

\begin{table*}[t]
\centering
\def\arraystretch{1.2}
\begin{tabular}{lccc}
\hline
\multirow{2}{*}{\textbf{Dataset}} & \textbf{Proportion of New Tokens} & \multicolumn{2}{c}{\textbf{Performance Compared to}} \\ \cline{3-4}
 & \textbf{(\%)} & $\textbf{IndoBERT}_{\textbf{BASE}}$ & $\textbf{IndoBERT}_{\textbf{LARGE}}$ \\ \hline
NusaX & 8.46 & \plus1.3 & \plus2.6 \\
NusaTranslation Emotion & 8.21 & \plus2.04 & \plus3.61 \\
NusaTranslation Sentiment & 6.83 & \plus1.83 & \plus2.14 \\
NusaParagraph Topic & 11.26 & \minus2.55 & \minus0.33 \\
NusaParagraph Rhetoric & 11.63 & \plus3.42 & \plus21.30 \\
NusaParagraph Emotion & 11.41 & \plus0.06 & \plus9.18 \\
\hline
\end{tabular}
\caption{\label{nusabert-propo-gains} Proportion of new tokens found only in the extended NusaBERT tokenizer compared with the performance gain of NusaBERT over IndoBERT for each dataset.}
\end{table*}

The official baseline result of NusaWrites aggregates the scores across all languages into a single mean score for each subtask \cite{cahyawijaya-etal-2023-nusawrites}. Fortunately, the individual raw results for each subtask and each language are available on the official NusaWrites repository\footnote{\url{https://github.com/IndoNLP/nusa-writes/}}, enabling us to thoroughly examine and compare per-language results. The aggregated baseline and NusaBERT results are shown in Table \ref{nusawrites-results}, while the detailed per-task and per-language results are shown in Appendix \ref{sec:nusawrites-complete-results}.

Like our results on NusaX, NusaBERT increases the average score on the two tasks of NusaTranslation. Specifically, $\text{NusaBERT}_{\text{BASE}}$ improves the NusaTranslation emotion classification score of IndoBERT from 52.59\% to 57.80\% (\plus5.21\%) and $\text{NusaBERT}_{\text{LARGE}}$ from 54.50\% to 61.40\% (\plus6.9\%). Further, on the sentiment analysis task, $\text{NusaBERT}_{\text{BASE}}$ improves the IndoBERT score from 75.24\% to 77.07\% (\plus1.83\%) and $\text{NusaBERT}_{\text{LARGE}}$ from 77.40\% to 79.54\% (\plus2.14\%). Overall, $\text{NusaBERT}_{\text{LARGE}}$ attained state-of-the-art results on both NusaTranslation tasks.

Notably, unlike NusaX, most languages of NusaTranslation are not found in the pre-training corpus of NusaBERT and are extremely low-resource. Nonetheless, based on the results alone, it seems that the introduction of additional new regional languages during the continued pre-training phase benefits the robustness of NusaBERT on these new languages as well, suggesting cross-lingual transferability. Similarly, NusaBERT’s results on languages that were included in the continued pre-training corpus like Javanese (\verb|jav|) and Minangkabau (\verb|min|) significantly improve that of IndoBERT. However, as noted by \citet{cahyawijaya-etal-2023-nusawrites}, NusaTranslation and NusaX share a similar source domain of mainly social media texts, therefore it is expected that our findings are parallel.

NusaParagraph, on the contrary, presents an even more challenging task by consisting of not only languages that are not found in our pre-training corpus but is also lexically more diverse and contains a remarkably higher ratio of local/colloquial words \cite{cahyawijaya-etal-2023-nusawrites}. Indeed, the gains of NusaBERT over IndoBERT are lackluster when evaluated on the NusaParagraph topic classification task. For instance, NusaBERT failed to improve the results of IndoBERT, dropping the result of $\text{IndoBERT}_{\text{BASE}}$ from 85.87\% to 83.32\% (\minus2.55\%), and for $\text{NusaBERT}_{\text{LARGE}}$, the result dropped from 85.41\% to 85.08\% (\minus0.33\%). Nevertheless, it still improved the IndoBERT results on both the rhetorical mode (\plus3.42\% $\text{NusaBERT}_{\text{BASE}}$, \plus21.3\% $\text{NusaBERT}_{\text{LARGE}}$) and emotion classification (\plus0.06\% $\text{NusaBERT}_{\text{BASE}}$, \plus9.18\% $\text{NusaBERT}_{\text{LARGE}}$) tasks. It is only on NusaParagraph rhetorical mode classification where $\text{NusaBERT}_{\text{LARGE}}$ gets state-of-the-art results.

Like the findings of \citet{cahyawijaya-etal-2023-nusawrites}, NusaBERT still fails to outperform classical machine learning baselines on languages that are highly distinct from Indonesian (\verb|ind|). We also note that NusaBERT was pre-trained on Wikipedia and common crawl corpora, which explains its effectiveness on and closeness to NusaX and NusaTranslation source domains, but not so for NusaParagraph. Thus, due to their high linguistic and lexical discrepancies, NusaBERT’s capabilities to exploit knowledge and cross-lingual transfer to these extremely low-resource languages remain largely ineffective. Further work is required to fully resolve these issues.

\subsection{Impact of New Tokens on Downstream Tasks}

\begin{table*}[t]
\centering
\def\arraystretch{1.2}
\begin{tabular}{lcccccc}
\hline
\textbf{Model} & \textbf{Original (ind)} & \textbf{eng} & \textbf{jav} & \textbf{msa} & \textbf{sun} & $\mu$ \\ \hline
\multicolumn{7}{c}{\textbf{EmoT}} \\ \hline
mBERT & 61.14 & 12.50 & 14.02 & 12.73 & 12.50 & 12.94 \\ 
$\text{XLM-R}_{\text{BASE}}$ & 72.88 & 10.98 & 13.94 & 13.18 & 12.50 & 12.65 \\ 
$\text{XLM-R}_{\text{LARGE}}$ & \underline{78.26} & 12.27 & 13.03 & 12.42 & 11.74 & 12.37 \\ 
$\text{IndoBERT}_{\text{BASE}}$ & 72.42 & 9.55 & 12.35 & \textbf{9.47} & 9.39 & \textbf{10.19} \\ 
$\text{IndoBERT}_{\text{LARGE}}$ & 75.53 & \textbf{9.24} & 12.12 & 10.23 & \textbf{9.32} & 10.23 \\ 
$\text{NusaBERT}_{\text{BASE}}$ & 75.23 & 14.09 & 14.77 & 13.64 & 13.64 & 14.03 \\ 
$\text{NusaBERT}_{\text{LARGE}}$ & 78.18 & 10.45 & \textbf{10.45} & 10.45 & 12.05 & 10.85 \\ \hline
\multicolumn{7}{c}{\textbf{SmSA}} \\ \hline
mBERT & 83.00 & 2.20 & 3.00 & 2.93 & 2.47 & 2.65 \\ 
$\text{XLM-R}_{\text{BASE}}$ & 91.53 & 3.40 & 3.80 & 4.27 & 4.27 & 3.94 \\ 
$\text{XLM-R}_{\text{LARGE}}$ & \underline{94.07} & 2.13 & 3.20 & 2.60 & 2.73 & 2.67 \\ 
$\text{IndoBERT}_{\text{BASE}}$ & 91.00 & 1.33 & 5.07 & 3.20 & 2.40 & 3.00 \\ 
$\text{IndoBERT}_{\text{LARGE}}$ & 94.20 & 2.47 & 4.13 & 4.00 & 2.20 & 3.20 \\ 
$\text{NusaBERT}_{\text{BASE}}$ & 91.00 & \textbf{0.60} & \textbf{2.80} & 2.40 & \textbf{1.80} & \textbf{1.90} \\ 
$\text{NusaBERT}_{\text{LARGE}}$ & 91.00 & 1.80 & 3.80 & \textbf{2.20} & 2.20 & 2.50 \\ \hline

\end{tabular}
\caption{\label{code-mixing-performance} Evaluation results on code-mixed downstream tasks, measured in delta accuracy with $R = 0.4$. Baseline results are obtained from \citet{adilazuarda-etal-2022-indorobusta}. The lowest delta accuracy on each task is \textbf{bolded} for clarity. The best-performing model on the originally Indonesian (\texttt{ind}) fine-tuning task has also been \underline{underlined}.}
\end{table*}

We investigated the impact of the new tokens on downstream tasks, especially noting that our extended tokenizer was additionally trained on the regional languages of Indonesia and that the IndoBERT tokenizer \cite{wilie-etal-2020-indonlu} might not be suitable for this purpose.

We modified the approach conducted by \citet{sriwirote2023phayathaibert}, where they calculated the proportion of unassimilated English words with respect to the number of total words in the downstream task. However, since we are unable to distinguish the regional languages’ words from the Indonesian words programmatically, we defined a new metric as follows:
\begin{equation}
    \text{Proportion of New Tokens} = \frac{\#\text{new tokens}}{\#\text{total tokens}}
\end{equation}

That is, we re-tokenized all downstream tasks’ texts using the extended NusaBERT tokenizer and calculated the percentage of new tokens with respect to the total number of tokens. This way, we can closely inspect and compare the relation between the newly introduced tokens and the gains of NusaBERT over IndoBERT. Table \ref{nusabert-propo-gains} shows the aforementioned results.

While the trend of the proportion of new tokens with the gains of NusaBERT over IndoBERT isn’t always linear, there is generally a correlation between the two -- parallel with the findings of \cite{sriwirote2023phayathaibert}. This, however, doesn’t apply to NusaParagraph topic classification where NusaBERT performed worse than IndoBERT. Despite these findings, the new tokens might not definitively be the only attributing factor to the improved results of NusaBERT (e.g. continued pre-training corpus), and further investigation is required.

\subsection{Code-mixing Robustness}

Although NusaBERT doesn’t directly address the issue of code-switching/code-mixing, we examined its code-mixing robustness by evaluating our models on IndoRobusta-Blend \cite{adilazuarda-etal-2022-indorobusta}. Following the suggested procedure, we took NusaBERT models which have been fine-tuned on the original Indonesian EmoT and SmSA datasets \cite{wilie-etal-2020-indonlu}, and conducted zero-shot inference on code-mixed versions of their respective test sets. To have a fair comparison with the official reported results, we similarly applied a perturbation ratio $R = 0.4$ and mixed English (\verb|eng|), Javanese (\verb|jav|), Malay (\verb|msa|), and Sundanese (\verb|sun|) as target L2 languages. We report the evaluation results in Table \ref{code-mixing-performance}. We also provided the full results in Appendix \ref{sec:indorobusta-complete-results}.

Interestingly, the robustness of NusaBERT depends highly on the downstream task being tested, similar to the findings of \citet{adilazuarda-etal-2022-indorobusta}. On sentiment analysis (SmSA), $\text{NusaBERT}_{\text{BASE}}$ is the most robust, significantly improving the robustness of $\text{IndoBERT}_{\text{BASE}}$. However, this doesn’t apply to emotion classification (EmoT) where $\text{NusaBERT}_{\text{LARGE}}$ is more robust than its $\text{NusaBERT}_{\text{BASE}}$. Further, both NusaBERT models are more prone to code-mixing on emotion classification compared to IndoBERT, but the opposite is true for sentiment analysis.

Additionally, parallel to what was conjectured by \citet{adilazuarda-etal-2022-indorobusta}, NusaBERT is generally more robust against Indonesian-English code-mixing. We agree with their suggestion that this stems from the source bias found in most online pre-training corpora that often mix these two languages. In the same light, Wikipedia texts that we pre-trained on also contain a high ratio of English loan words \cite{cahyawijaya-etal-2023-nusawrites}, thereby explaining these findings.

\section{Conclusion}
In this study, we introduced NusaBERT, a multilingual language model specifically tailored to the linguistic diversity of Indonesia. Basing our model on IndoBERT, we applied vocabulary expansion and continued pre-training on a multilingual corpus that introduces the regional languages of Indonesia. NusaBERT achieves state-of-the-art results when evaluated on Indonesian and multilingual NLU benchmarks such as IndoNLU, NusaX, and NusaWrites. These findings highlight the effectiveness of our proposed approach in enhancing the multilingual and multicultural capabilities of IndoBERT to address Indonesia's unique linguistic framework. We also discussed several limitations of NusaBERT and how to potentially resolve them. We hope NusaBERT will enable further research in the under-represented languages of Indonesia.

\section*{Limitations}

\subsection{Code-switching}

NusaBERT demonstrates proficiency in handling low-resource languages while surpassing or remaining competitive with monolingual models on downstream tasks. Despite this efficacy, it has yet to address the intricate challenge of intra-sentential code-switching. While the issue of code-switching is not explicitly tackled in the context of NusaBERT, results in Table \ref{code-mixing-performance} indicate potential room for improvements that can be done to enhance NusaBERT's performance in handling code-switching scenarios. Moreover, it is important to mention that the language model’s performance on IndoRobusta-Blend does not definitively represent its robustness against code-switching as it uses synthetically generated code-mixed examples instead of human-curated code-mixed data, and is limited to only four L2 languages. Having an expert-curated code-mixing benchmark would be valuable for future evaluations.

To tackle code-switching adversarial attacks, \citet{adilazuarda-etal-2022-indorobusta} proposed a code-mixing adversarial training technique called IndoRobusta-Shot that suggests three different fine-tuning techniques: code-mixed-only tuning, two-step tuning, and joint training. Among the three examined methods, joint training shows the best results which implies that training code-mixed data with monolingual data increases the robustness of language models while maintaining its monolingual downstream capabilities. 

\subsection{Adapting NusaBERT to New Languages}

In our study, we introduced a multilingual language model designed for Indonesian and its 12 regional languages. Although 12 languages is considerably a large number, it is considered comparatively modest compared to Indonesia’s boasting rich linguistic landscape with over 700 languages and dialects. This arises from the significant difference in the amount of available text corpus of regional languages and the lack of quality data.

Several endeavors have successfully extended new languages to a base language model. For example, the BLOOM language model \cite{workshop2023bloom}, a comprehensive multilingual language model trained on 46 languages, effectively extended its applicability to 8 previously unseen languages \cite{yong-etal-2023-bloom} through continued pretraining, implementation of language adapters \cite{pfeiffer-etal-2020-mad}, and parameter-efficient finetuning techniques \cite{liu2022few}. These strategies facilitated the inclusion of new languages while preserving existing capabilities and mitigating catastrophic forgetting. Despite the demonstrated feasibility of extending language models to existing language models, the data on these new languages are abundant in comparison to Indonesian regional languages.

A recent approach proposed by \citet{wang-etal-2022-expanding} seeks to leverage bilingual lexicons which are widely available even for extremely low-resource languages. We can thereby potentially generate synthetic low-resource language texts by translating from Indonesian texts using these lexicons. This approach, coupled with gold few-texts of the target language, if available, is one way to possibly extend NusaBERT to extremely low-resource languages where resources are highly scarce.

\subsection{Corpus Domain Diversity}

One significant limitation in our study is the lack of corpus domain diversity, particularly evident in the performance discrepancies between NusaParagraph and the other tasks (NusaX and NusaTranslation). The underpinning challenge with NusaParagraph, which diverges from the social media domain to include paragraph writing by human annotators, is its richer cultural and lexical diversity, indicative of the nuanced and colloquial language use in very low-resource and linguistically distinct local languages \cite{cahyawijaya-etal-2023-nusawrites}. This complexity is inherently difficult for models like NusaBERT, which, despite their robustness, are pre-trained predominantly on social media texts and online documents similar to the datasets used for NusaX and NusaTranslation.

Despite the apparent scarcity of directly applicable, culturally rich, and linguistically aligned corpora for very low-resource local languages, there exists an opportunity to leverage alternative texts during model pre-training. For instance, texts such as the Bible, which are often translated into numerous languages, including many under-represented ones, could provide a valuable resource \cite{wongso-etal-2023-many}. These texts offer a range of linguistic structures and vocabularies that, while not entirely reflective of colloquial use, could serve as a foundational step towards bridging the gap in language representation. This approach underscores the necessity for creative solutions in the absence of conventional data sources, aiming to enhance the model's performance across a wider array of linguistic contexts.

This strategy invites further research to not only incorporate existing texts from under-represented languages into pre-training processes but also to innovate methods such as leveraging and exploring the use of non-text data. Specifically, transcribing conversation audio through speech recognition, especially for local Indonesian languages that are rarely ever written \cite{aji-etal-2022-one}, presents a novel avenue to enrich the language’s resources. This approach can capture the authentic linguistic nuances and cultural richness of spoken language, offering a more comprehensive representation of these languages \cite{BESACIER201485}.

This direction not only underscores the ongoing effort to fully leverage the linguistic diversity of Indonesia and similar regions but also aims to expand the applicability and inclusivity of language models by incorporating the rich, oral traditions of under-represented communities into the digital linguistic landscape.


\bibliography{anthology,custom}
\bibliographystyle{acl_natbib}

\appendix

\section{Languages Studied}
\label{sec:language-studied}

The list of languages and dialects included in this study and their statistics are shown in Table \ref{languages-studied}.

\begin{table}[ht]
\small
\centering
\setlength{\tabcolsep}{0.8\tabcolsep}
\def\arraystretch{1.2}
\begin{tabular}{llc}
\hline
\textbf{Language} & \textbf{Primary Region} & \textbf{\#speakers} \\
\hline
Acehnese (\verb|ace|) & Aceh & 2,840,000 \\
Ambon (\verb|abs|) & Maluku & 1,650,900 \\
Balinese (\verb|ban|) & Bali & 3,300,000 \\
Banjarese (\verb|bjn|) & Kalimantan & 3,650,000 \\
Banyumasan (\verb|jav|) & Banyumasan & N/A \\
Batak (\verb|btk|) & North Sumatra & 3,320,000\textsuperscript{*} \\
Betawi (\verb|bew|) & Banten, Jakarta & 5,000,000 \\
Bima (\verb|bhp|) & Sumbawa & 500,000 \\
Buginese (\verb|bug|) & South Sulawesi & 4,370,000 \\
Gorontalo (\verb|gor|) & Gorontalo & 505,000 \\
Indonesian (\verb|ind|) & Indonesia & 198,000,000 \\
Javanese (\verb|jav|) & Java & 68,200,000 \\
Madurese (\verb|mad|) & Madura & 7,790,000 \\
Makassarese (\verb|mak|) & South Sulawesi & 2,110,000 \\
Malay (\verb|msa|) & Malaysia & 82,285,706 \\
Minangkabau (\verb|min|) & West Sumatra & 4,880,000 \\
Musi (\verb|mui|) & South Sumatra & 3,105,000 \\
Ngaju (\verb|nij|) & Central Kalimantan & 890,000 \\
Nias (\verb|nia|) & Nias & 867,000 \\
Rejang (\verb|rej|) & Bengkulu & 350,000 \\
Sundanese (\verb|sun|) & West Java & 32,400,000 \\
Tetum (\verb|tet|) & East Timor & 91,200 \\
Toba Batak (\verb|bbc|) & North Sumatra & 1,610,000 \\
\hline
\end{tabular}
\caption{\label{languages-studied} Statistics of languages included in this study, with data obtained from \citet{ethnologue} and \citet{bps2010kewarganegaraan}\textsuperscript{*}.}
\end{table}

\section{NusaWrites Evaluation Results}
\label{sec:nusawrites-complete-results}

We included the non-aggregated, per-task, and per-language evaluation results of NusaWrites. NusaTranslation results are shown in Table \ref{nusatranslation-emot} and Table \ref{nusatranslation-senti}. NusaParagraph results are shown in Table \ref{nusaparagraph-topic}, Table \ref{nusaparagraph-rhetoric}, and Table \ref{nusaparagraph-emot}.

\begin{table*}[ht]
\small
\centering
\setlength{\tabcolsep}{0.5\tabcolsep}
\def\arraystretch{1.2}
\begin{tabular}{lcccccccccccc}
\hline
\multicolumn{13}{c}{\textbf{NusaTranslation EmoT}} \\
\hline
\textbf{Model} & \textbf{abs} & \textbf{bew} & \textbf{bhp} & \textbf{btk} & \textbf{jav} & \textbf{mad} & \textbf{mak} & \textbf{min} & \textbf{mui} & \textbf{rej} & \textbf{sun} & $\mu$ \\ \hline
Logistic Regression (Bag of Words) & 46.77 & 62.31 & 44.62 & 59.38 & 60.66 & 58.05 & 55.63 & 61.73 & 45.33 & 45.61 & 62.90 & \multirow{2}{*}{56.18} \\
Logistic Regression (TF-IDF) & \textbf{51.20} & 63.59 & \textbf{50.06} & 61.25 & 61.47 & 60.42 & 56.39 & 63.94 & 50.98 & 50.61 & 62.99 &  \\ \hline
Naive Bayes (Bag of Words) & 48.16 & 59.76 & 48.02 & 57.12 & 58.39 & 55.22 & 54.93 & 61.41 & 51.49 & 47.53 & 61.32 & \multirow{2}{*}{52.71} \\
Naive Bayes (TF-IDF) & 49.95 & 55.54 & 40.12 & 54.64 & 54.85 & 52.76 & 52.03 & 56.61 & 48.93 & 32.87 & 57.86 &  \\ \hline
SVM (Bag of Words) & 44.56 & 61.30 & 43.59 & 58.43 & 58.97 & 55.97 & 52.60 & 61.02 & 48.80 & 41.81 & 60.58 & \multirow{2}{*}{55.08} \\
SVM (TF-IDF) & 48.23 & 61.74 & 48.68 & 61.02 & 63.34 & 59.43 & 58.09 & 62.34 & 51.40 & 48.27 & 61.58 &  \\ \hline
mBERT & 26.05 & 59.75 & 12.65 & 59.28 & 62.80 & 57.30 & 54.92 & 61.50 & 16.48 & 12.24 & 62.49 & 44.13 \\
$\text{XLM-R}_{\text{BASE}}$ & 35.79 & 63.54 & 12.44 & 59.95 & 62.86 & 59.87 & 60.54 & 63.39 & 13.94 & 19.75 & 65.14 & 47.02 \\
$\text{XLM-R}_{\text{LARGE}}$ & 49.58 & 70.43 & 8.53 & \textbf{65.83} & 68.70 & 61.27 & 58.85 & \textbf{70.84} & 55.83 & 23.12 & 70.24 & 54.84 \\ \hline
$\text{IndoLEM IndoBERT}_{\text{BASE}}$ & 35.03 & 67.86 & 25.40 & 59.86 & 64.47 & 59.40 & 58.23 & 61.48 & 45.00 & 39.20 & 62.56 & 52.59 \\ 
$\text{IndoNLU IndoBERT}_{\text{BASE}}$ & 41.04 & 66.61 & 32.13 & 62.81 & 66.91 & 61.52 & \textbf{61.81} & 67.95 & 42.78 & 33.54 & 62.38 & 54.50 \\ 
$\text{IndoNLU IndoBERT}_{\text{LARGE}}$ & 48.54 & 72.55 & 28.43 & 63.09 & 69.34 & 61.84 & 60.48 & 67.55 & 53.22 & 40.19 & \textbf{70.53} & 57.80 \\ \hline
$\text{NusaBERT}_{\text{BASE}}$ & 45.21 & 66.09 & 39.03 & 61.72 & 67.41 & 61.10 & 60.54 & 67.11 & 50.98 & 37.36 & 65.34 & 56.54 \\ 
$\text{NusaBERT}_{\text{LARGE}}$ & 47.75 & \textbf{73.68} & 36.31 & 62.87 & \textbf{73.63} & \textbf{65.48} & 60.58 & 70.27 & \textbf{60.06} & \textbf{54.47} & 70.34 & \textbf{61.40} \\ \hline
\end{tabular}
\caption{\label{nusatranslation-emot} Evaluation results of baseline models and NusaBERT on the NusaTranslation emotion classification task, measured in macro-F1 (\%). Baseline results are obtained from \citet{cahyawijaya-etal-2023-nusawrites}. The best performance on each task is \textbf{bolded} for clarity.}
\end{table*}

\begin{table*}[ht]
\small
\centering
\setlength{\tabcolsep}{0.5\tabcolsep}
\def\arraystretch{1.2}
\begin{tabular}{lcccccccccccc}
\hline
\multicolumn{13}{c}{\textbf{NusaTranslation Senti}} \\
\hline
\textbf{Model} & \textbf{abs} & \textbf{bew} & \textbf{bhp} & \textbf{btk} & \textbf{jav} & \textbf{mad} & \textbf{mak} & \textbf{min} & \textbf{mui} & \textbf{rej} & \textbf{sun} & $\mu$ \\ \hline
Logistic Regression (Bag of Words) & 69.23 & 81.88 & 41.86 & 79.13 & 81.87 & 81.48 & 78.39 & 82.53 & 70.35 & 60.79 & 84.43 & \multirow{2}{*}{74.96} \\
Logistic Regression (TF-IDF) & 69.50 & 81.04 & \textbf{70.10} & 79.67 & 77.85 & 74.50 & 78.27 & 82.18 & 72.31 & 68.00 & 83.73 &  \\ \hline
Naive Bayes (Bag of Words) & 69.67 & 79.12 & 69.36 & 78.05 & 79.88 & 78.38 & 76.77 & 80.10 & 72.20 & 69.05 & 80.51 & \multirow{2}{*}{74.89} \\
Naive Bayes (TF-IDF) & 67.71 & 77.03 & 64.51 & 76.56 & 75.71 & 77.70 & 76.41 & 80.11 & 71.41 & 66.90 & 80.34 &  \\ \hline
SVM (Bag of Words) & 69.87 & 81.94 & 69.89 & 79.77 & 78.18 & 80.44 & 79.25 & 82.68 & 68.02 & 66.45 & 84.21 & \multirow{2}{*}{76.04} \\
SVM (TF-IDF) & 70.28 & 82.26 & 68.94 & 76.20 & 78.16 & 75.28 & 77.67 & 81.66 & 72.20 & 66.36 & 83.10 &  \\ \hline
mBERT & 67.47 & 79.56 & 41.86 & 72.81 & 80.55 & 76.44 & 69.08 & 79.43 & 64.07 & 46.03 & 78.56 & 68.71 \\
$\text{XLM-R}_{\text{BASE}}$ & 67.28 & 85.11 & 41.86 & 77.22 & 79.73 & 78.40 & 75.90 & 83.39 & 40.90 & 40.97 & 84.08 & 68.62 \\
$\text{XLM-R}_{\text{LARGE}}$ & \textbf{72.55} & 86.54 & 65.52 & 80.62 & 86.13 & 78.58 & \textbf{81.86} & 86.04 & \textbf{78.80} & 65.18 & \textbf{87.87} & 79.06 \\ \hline
$\text{IndoLEM IndoBERT}_{\text{BASE}}$ & 59.39 & 81.57 & 44.66 & 74.50 & 81.89 & 72.28 & 66.12 & 80.95 & 65.52 & 51.25 & 81.74 & 69.08 \\
$\text{IndoNLU IndoBERT}_{\text{BASE}}$ & 70.45 & 86.09 & 62.80 & 72.64 & 84.34 & 75.16 & 76.80 & 82.62 & 71.32 & 66.59 & 78.82 & 75.24 \\
$\text{IndoNLU IndoBERT}_{\text{LARGE}}$ & 72.16 & 87.92 & 59.91 & 78.39 & 81.61 & 79.84 & 78.96 & 81.99 & 75.98 & 68.79 & 85.83 & 77.40 \\ \hline
$\text{NusaBERT}_{\text{BASE}}$ & 70.71 & 86.02 & 63.72 & 80.63 & 84.04 & 80.47 & 80.73 & 84.75 & 66.14 & 64.80 & 85.74 & 77.07 \\
$\text{NusaBERT}_{\text{LARGE}}$ & 68.94 & \textbf{90.11} & 66.46 & \textbf{83.09} & \textbf{86.71} & \textbf{83.66} & 81.35 & \textbf{86.42} & 70.66 & \textbf{69.74} & 87.83 & \textbf{79.54} \\ \hline
\end{tabular}
\caption{\label{nusatranslation-senti} Evaluation results of baseline models and NusaBERT on the NusaTranslation sentiment analysis task, measured in macro-F1 (\%). Baseline results are obtained from \citet{cahyawijaya-etal-2023-nusawrites}. The best performance on each task is \textbf{bolded} for clarity.}
\end{table*}

\begin{table*}[ht]
\small
\centering
\setlength{\tabcolsep}{0.7\tabcolsep}
\def\arraystretch{1.2}
\begin{tabular}{lccccccccccc}
\hline
\multicolumn{12}{c}{\textbf{NusaParagraph Topic}} \\ \hline
\textbf{Model} & \textbf{bew} & \textbf{btk} & \textbf{bug} & \textbf{jav} & \textbf{mad} & \textbf{mak} & \textbf{min} & \textbf{mui} & \textbf{rej} & \textbf{sun} & $\mu$ \\ \hline
Logistic Regression (Bag of Words) & 90.20 & 88.95 & 68.87 & 90.65 & 88.87 & 87.50 & 90.70 & 85.71 & 82.22 & 89.67 & \multirow{2}{*}{\textbf{87.67}} \\
Logistic Regression (TF-IDF) & 92.63 & \textbf{91.09} & \textbf{73.92} & 91.49 & \textbf{92.32} & 91.21 & 92.10 & 88.02 & 86.39 & 90.87 &  \\ \hline
Naive Bayes (Bag of Words) & 87.72 & 84.55 & 62.88 & 87.32 & 82.40 & 89.27 & 90.64 & 86.21 & \textbf{88.09} & 89.45 & \multirow{2}{*}{85.06} \\
Naive Bayes (TF-IDF) & 89.11 & 85.38 & 60.06 & 89.55 & 83.44 & 90.26 & 89.96 & \textbf{88.20} & 86.58 & 90.10 &  \\ \hline
SVM (Bag of Words) & 89.48 & 85.59 & 61.46 & 87.79 & 86.49 & 84.85 & 89.55 & 82.51 & 78.36 & 88.28 & \multirow{2}{*}{85.86} \\
SVM (TF-IDF) & 91.76 & 90.25 & 73.57 & 90.64 & 90.61 & \textbf{91.34} & \textbf{92.56} & 86.06 & 84.88 & 91.19 &  \\ \hline
mBERT & 89.22 & 86.66 & 43.26 & 87.41 & 77.40 & 84.61 & 88.75 & 83.30 & 9.54 & 88.00 & 73.82 \\
$\text{XLM-R}_{\text{BASE}}$ & 90.11 & 86.84 & 46.11 & 89.82 & 83.59 & 84.22 & 88.19 & 3.45 & 54.23 & 90.26 & 71.68 \\
$\text{XLM-R}_{\text{LARGE}}$ & 92.33 & 85.75 & 43.18 & 91.07 & 85.81 & 85.60 & 89.06 & 85.69 & 81.04 & 91.00 & 83.05 \\ \hline
$\text{IndoLEM IndoBERT}_{\text{BASE}}$ & 91.74 & 87.23 & 61.53 & 90.52 & 86.50 & 87.96 & 90.82 & 85.00 & 78.77 & 88.59 & 84.87 \\ 
$\text{IndoNLU IndoBERT}_{\text{BASE}}$ & 91.64 & 87.26 & 67.72 & 90.59 & 85.00 & 85.30 & 90.50 & 86.52 & 85.74 & 88.43 & 85.87 \\ 
$\text{IndoNLU IndoBERT}_{\text{LARGE}}$ & 92.17 & 85.95 & 66.79 & 90.05 & 87.11 & 87.11 & 91.30 & 86.16 & 78.06 & 89.39 & 85.41 \\ \hline
$\text{NusaBERT}_{\text{BASE}}$ & 91.81 & 87.27 & 52.45 & 91.45 & 87.48 & 87.61 & 91.97 & 83.05 & 77.57 & 91.03 & 83.32 \\ 
$\text{NusaBERT}_{\text{LARGE}}$ & \textbf{93.18} & 87.20 & 60.97 & \textbf{93.44} & 85.80 & 88.93 & 92.25 & 87.15 & 77.48 & \textbf{92.48} & 85.08 \\ \hline
\end{tabular}
\caption{\label{nusaparagraph-topic} Evaluation results of baseline models and NusaBERT on the NusaParagraph topic classification task, measured in macro-F1 (\%). Baseline results are obtained from \citet{cahyawijaya-etal-2023-nusawrites}. The best performance on each task is \textbf{bolded} for clarity.}
\end{table*}

\begin{table*}[ht]
\small
\centering
\setlength{\tabcolsep}{0.7\tabcolsep}
\def\arraystretch{1.2}
\begin{tabular}{lccccccccccc}
\hline
\multicolumn{12}{c}{\textbf{NusaParagraph Rhetoric}} \\
\hline
\textbf{Model} & \textbf{bew} & \textbf{btk} & \textbf{bug} & \textbf{jav} & \textbf{mad} & \textbf{mak} & \textbf{min} & \textbf{mui} & \textbf{rej} & \textbf{sun} & $\mu$ \\ \hline
Logistic Regression (Bag of Words) & 39.40 & 33.85 & 61.77 & 64.52 & 47.97 & 23.87 & 59.09 & 53.82 & 28.46 & 49.39 & \multirow{2}{*}{45.21} \\
Logistic Regression (TF-IDF) & 40.10 & 33.10 & 57.11 & 64.85 & 48.56 & 24.08 & 57.68 & 44.67 & 22.70 & 49.28 &  \\ \hline
Naive Bayes (Bag of Words) & 37.78 & 28.23 & 51.29 & 56.94 & 42.62 & 22.78 & 46.92 & 35.55 & 20.95 & 44.79 & \multirow{2}{*}{37.73} \\
Naive Bayes (TF-IDF) & 36.79 & 26.06 & 44.02 & 53.68 & 42.89 & 22.98 & 44.67 & 32.65 & 20.72 & 42.22 &  \\ \hline
SVM (Bag of Words) & 41.51 & 32.04 & 60.55 & 67.12 & 48.21 & 23.25 & 59.50 & 50.09 & 31.76 & 49.98 & \multirow{2}{*}{45.44} \\
SVM (TF-IDF) & 40.76 & 32.60 & 57.29 & 65.07 & 48.28 & 22.22 & 57.79 & 45.51 & 26.13 & 49.18 &  \\ \hline
mBERT & 43.21 & 24.92 & 70.26 & 74.29 & 53.02 & 17.52 & 67.37 & 61.67 & 32.85 & 54.94 & 50.01 \\
$\text{XLM-R}_{\text{BASE}}$ & 48.75 & 23.08 & 70.03 & 78.04 & 52.09 & 8.28 & 68.60 & 61.80 & 22.83 & 58.17 & 49.17 \\
$\text{XLM-R}_{\text{LARGE}}$ & 50.52 & 29.07 & 68.62 & \textbf{78.43} & 53.78 & 16.47 & \textbf{72.80} & \textbf{64.81} & 21.91 & \textbf{59.29} & 51.57 \\ \hline
$\text{IndoLEM IndoBERT}_{\text{BASE}}$ & 48.73 & 31.48 & 65.72 & 74.23 & 51.80 & \textbf{24.87} & 68.66 & 64.07 & \textbf{36.45} & 53.32 & 51.93 \\
$\text{IndoNLU IndoBERT}_{\text{BASE}}$ & 47.40 & 29.14 & 53.40 & 69.24 & 51.59 & 20.42 & 64.75 & 57.11 & 34.07 & 52.11 & 47.92 \\
$\text{IndoNLU IndoBERT}_{\text{LARGE}}$ & 6.64 & 7.62 & 6.80 & 73.59 & 48.13 & 11.80 & 66.32 & 17.37 & 25.38 & 53.91 & 31.76 \\ \hline
$\text{NusaBERT}_{\text{BASE}}$ & 48.76 & \textbf{34.61} & 60.05 & 74.74 & 52.43 & 24.73 & 68.02 & 60.83 & 31.57 & 57.65 & 51.34 \\
$\text{NusaBERT}_{\text{LARGE}}$ & \textbf{50.25} & 33.38 & \textbf{72.52} & 78.23 & \textbf{54.47} & 18.38 & 69.18 & 64.71 & 32.55 & 56.89 & \textbf{53.06} \\ \hline
\end{tabular}
\caption{\label{nusaparagraph-rhetoric} Evaluation results of baseline models and NusaBERT on the NusaParagraph rhetoric mode classification task, measured in macro-F1 (\%). Baseline results are obtained from \citet{cahyawijaya-etal-2023-nusawrites}. The best performance on each task is \textbf{bolded} for clarity.}
\end{table*}

\begin{table*}[ht]
\small
\centering
\setlength{\tabcolsep}{0.7\tabcolsep}
\def\arraystretch{1.2}
\begin{tabular}{lccccccccccc}
\hline
\multicolumn{12}{c}{\textbf{NusaParagraph EmoT}} \\ \hline
\textbf{Model} & \textbf{btk} & \textbf{bew} & \textbf{bug} & \textbf{jav} & \textbf{mad} & \textbf{mak} & \textbf{min} & \textbf{mui} & \textbf{rej} & \textbf{sun} & $\mu$ \\ \hline
Logistic Regression (Bag of Words) & 82.03 & 78.33 & 55.89 & 84.77 & \textbf{75.20} & 72.90 & 89.52 & 71.82 & 72.43 & 83.09 & \multirow{2}{*}{\textbf{78.23}} \\
Logistic Regression (TF-IDF) & 84.52 & \textbf{83.68} & 64.53 & \textbf{88.04} & 69.87 & \textbf{79.55} & \textbf{91.01} & 71.84 & \textbf{80.67} & 84.93 &  \\ \hline
Naive Bayes (Bag of Words) & 78.28 & 71.84 & \textbf{68.08} & 81.37 & 66.53 & 71.43 & 87.07 & 75.39 & 75.72 & 79.42 & \multirow{2}{*}{75.51} \\
Naive Bayes (TF-IDF) & 77.97 & 75.06 & 62.92 & 83.15 & 68.27 & 75.80 & 85.98 & 71.34 & 75.95 & 78.57 &  \\ \hline
SVM (Bag of Words) & 80.45 & 76.61 & 53.76 & 82.26 & 73.26 & 71.90 & 87.05 & 69.06 & 69.42 & 81.36 & \multirow{2}{*}{76.36} \\
SVM (TF-IDF) & 84.51 & 82.50 & 65.27 & 86.96 & 70.64 & 78.74 & 89.09 & 71.85 & 66.66 & 85.82 &  \\ \hline
mBERT & 80.60 & 65.35 & 26.49 & 78.90 & 58.84 & 58.40 & 82.56 & 63.66 & 39.97 & 76.74 & 63.15 \\ 
$\text{XLM-R}_{\text{BASE}}$ & 81.38 & 64.15 & 11.17 & 83.28 & 53.25 & 51.98 & 83.79 & 61.12 & 22.38 & 78.94 & 59.14 \\
$\text{XLM-R}_{\text{LARGE}}$ & \textbf{86.92} & 70.39 & 30.84 & 85.50 & 57.31 & 60.45 & 84.40 & 78.59 & 32.11 & \textbf{87.74} & 67.43 \\ \hline
$\text{IndoLEM IndoBERT}_{\text{BASE}}$ & 86.59 & 66.80 & 36.81 & 84.58 & 54.75 & 59.39 & 82.99 & 63.76 & 57.31 & 76.39 & 66.94 \\
$\text{IndoNLU IndoBERT}_{\text{BASE}}$ & 83.04 & 67.59 & 31.83 & 82.01 & 59.35 & 62.00 & 84.08 & 74.60 & 49.40 & 77.27 & 67.12 \\
$\text{IndoNLU IndoBERT}_{\text{LARGE}}$ & 85.49 & 71.92 & 27.88 & 84.52 & 43.55 & 66.51 & 81.75 & 74.87 & 13.06 & 76.89 & 62.64 \\ \hline
$\text{NusaBERT}_{\text{BASE}}$ & 84.44 & 74.19 & 36.44 & 84.18 & 59.16 & 66.70 & 85.61 & 66.37 & 36.54 & 78.13 & 67.18 \\
$\text{NusaBERT}_{\text{LARGE}}$ & 86.57 & 74.06 & 44.94 & 85.86 & 72.31 & 73.14 & 86.83 & \textbf{82.96} & 30.19 & 81.36 & 71.82 \\ \hline
\end{tabular}
\caption{\label{nusaparagraph-emot} Evaluation results of baseline models and NusaBERT on the NusaParagraph emotion classification task, measured in macro-F1 (\%). Baseline results are obtained from \citet{cahyawijaya-etal-2023-nusawrites}. The best performance on each task is \textbf{bolded} for clarity.}
\end{table*}

\section{IndoRobusta Evaluation Results}
\label{sec:indorobusta-complete-results}

The evaluation results of baseline models and NusaBERT on IndoRobusta-Blend are shown in Table \ref{indorobusta-complete-results}.

\begin{table*}[ht]
\centering
\def\arraystretch{1.2}
\begin{tabular}{lcccccc}
\hline
\textbf{Model} & \textbf{Original (ind)} & \textbf{eng} & \textbf{jav} & \textbf{msa} & \textbf{sun} & $\mu$ \\ \hline
\multicolumn{7}{c}{\textbf{EmoT}} \\ \hline
$\text{IndoBERT}_{\text{BASE}}$ & 72.42 & 62.87 & 60.07 & 62.95 & 63.03 & 64.27 \\
$\text{IndoBERT}_{\text{LARGE}}$ & 75.53 & 66.29 & 63.41 & 65.30 & 66.21 & 67.35 \\ \hline
mBERT & 61.14 & 48.64 & 47.12 & 48.41 & 48.64 & 50.79 \\
$\text{XLM-R}_{\text{BASE}}$ & 72.88 & 61.90 & 58.94 & 59.70 & 60.38 & 62.76 \\
$\text{XLM-R}_{\text{LARGE}}$ & \textbf{78.26} & 65.99 & 65.23 & 65.84 & \textbf{66.52} & 68.37 \\ \hline
$\text{NusaBERT}_{\text{BASE}}$ & 75.23 & 61.14 & 60.45 & 61.59 & 61.59 & 64.00 \\
$\text{NusaBERT}_{\text{LARGE}}$ & 78.18 & \textbf{67.73} & \textbf{67.73} & \textbf{67.73} & 66.14 & \textbf{69.50} \\ \hline
\multicolumn{7}{c}{\textbf{SmSA}} \\ \hline
$\text{IndoBERT}_{\text{BASE}}$ & 91.00 & 89.67 & 85.93 & 87.80 & 88.60 & 88.60 \\
$\text{IndoBERT}_{\text{LARGE}}$ & \textbf{94.20} & 91.73 & 90.07 & 90.20 & \textbf{92.00} & 91.64 \\ \hline
mBERT & 83.00 & 80.80 & 80.00 & 80.07 & 80.53 & 80.88 \\
$\text{XLM-R}_{\text{BASE}}$ & 91.53 & 88.13 & 87.73 & 87.26 & 87.26 & 88.38 \\
$\text{XLM-R}_{\text{LARGE}}$ & 94.07 & \textbf{91.94} & \textbf{90.87} & \textbf{91.47} & 91.34 & \textbf{91.94} \\ \hline
$\text{NusaBERT}_{\text{BASE}}$ & 91.00 & 90.40 & 88.20 & 88.60 & 89.20 & 89.48 \\
$\text{NusaBERT}_{\text{LARGE}}$ & 91.00 & 89.20 & 87.20 & 88.80 & 88.80 & 89.00 \\ \hline
\end{tabular}
\caption{\label{indorobusta-complete-results} Code-mixing robustness evaluation results of baseline models and NusaBERT on IndoRobusta-Blend, measured in accuracy (\%). Baseline results are inferred from the delta accuracies reported by \citet{adilazuarda-etal-2022-indorobusta}. The best performance on each task is \textbf{bolded} for clarity.}
\end{table*}

\end{document}